\documentclass[runningheads]{llncs}
\usepackage[dvipdfmx]{graphicx}
\usepackage{bm}
\usepackage{amsmath}
\usepackage{amsfonts}
\usepackage{amssymb}

\usepackage{color}
\newcommand{\letter}[1]{`{\tt #1}'}
\newcommand{\letterp}[1]{`{\tt #1}.'}

\makeatletter
\newcommand{\figcaption}[1]{\def\@captype{figure}\caption{#1}}
\newcommand{\tblcaption}[1]{\def\@captype{table}\caption{#1}}
\makeatother
\begin{document}
\title{Cross-Domain Image Conversion by CycleDM}
\titlerunning{Cross-Domain Image Conversion by CycleDM}
% If the paper title is too long for the running head, you can set
% an abbreviated paper title here
%
% \author{First Author\inst{1}\orcidID{0000-1111-2222-3333} \and
% Second Author\inst{2,3}\orcidID{1111-2222-3333-4444} \and
% Third Author\inst{3}\orcidID{2222--3333-4444-5555}}

\author{Sho Shimotsumagari\and
Shumpei Takezaki\and
Daichi Haraguchi\orcidID{0000-0002-3109-9053}\and
Seiichi Uchida\orcidID{0000-0001-8592-7566}}
\authorrunning{S. Shimotsumagari et al.}

\institute{Kyushu University, Fukuoka, Japan\\
\email{\{sho.shimotsumagari, shumpei.takezaki, uchida\}@human.ait.kyushu-u.ac.jp}}

\maketitle              % typeset the header of the contribution
\begin{abstract} 
% The abstract should briefly summarize the contents of the paper in
% 15--250 words.
The purpose of this paper is to enable the conversion between machine-printed character images (i.e., font images) and handwritten character images through machine learning. For this purpose, we propose a novel unpaired image-to-image domain conversion method, CycleDM,  which incorporates the concept of CycleGAN into the diffusion model. Specifically, CycleDM has two internal conversion models that bridge the denoising processes of two image domains. These conversion models are efficiently trained without explicit correspondence between the domains. By applying machine-printed and handwritten character images to the two modalities, CycleDM realizes the conversion between them. Our experiments for evaluating the converted images quantitatively and qualitatively found that ours performs better 
than other comparable approaches.
% 7:10 SU
\end{abstract}
\begin{figure}[t]
    \centering
    \includegraphics[width=0.5\linewidth]{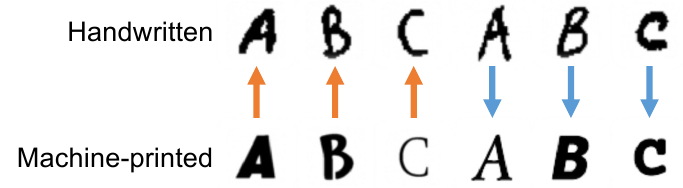}\\[-3mm]
    \caption{Cross-domain conversion task between machine-printed and handwritten character images. The converted image should resemble the original to some degree.}
    \label{fig:conversion-task}
\end{figure}%

% \begin{figure}[t]
%     \centering
%     \vspace{5cm}
%     \includegraphics[width=.9\linewidth]{img/_method/img2img-method.pdf}\\[-3mm]
%     \caption{Cross-domain conversion task between machine-printed and handwritten character images.}
%     \label{fig:img2img-method}
% \end{figure}%

% ===================================================================
\section{Introduction\label{sec:intro}}
% ===================================================================
We consider a domain conversion task between machine-printed character images (i.e., font images) and handwritten character images. Fig.~\ref{fig:conversion-task} shows examples of this conversion task; 
for example, a machine-printed `A' should be converted to a {\em similar} handwritten `A,' and vice versa. Despite sharing the same character symbols (such as `A'), printed and handwritten characters exhibit significant differences in shape variations.
Machine-printed characters often show ornamental elements like serifs and changes in stroke width, whereas handwritten characters do not. On the other hand, handwritten characters often show variations by the fluctuations of the starting and ending positions of strokes or substantial shape changes by cursive writing, whereas machine-printed characters do not. Consequently, despite representing the same characters, their two domains are far from identical, making their mutual conversion a very challenging task.\par 
% 7:19 SU
%
Our domain conversion task is motivated from four perspectives. The first motivation is a technical interest in tackling a hard domain shift problem. As previously mentioned, the domain gap between handwritten and printed characters seems large, even within the same character class. Moreover, differences in character classes are often much smaller than the domain gap; for example, the difference between \letter{I} and \letter{J} is often smaller than the difference between a handwritten \letter{I} and a printed \letterp{I}  
Note that character images have been typical targets of domain ``adaptation''; especially scene digit images (e.g., Street View House Number, SVHN) and handwritten digit images (e.g., MNIST) are frequently employed as two domains~\cite{murez2018image,Tzeng_2017_CVPR}.
However, they are employed in the domain adaptation for better character-class recognition systems (i.e., OCR) rather than domain ``conversion,'' and therefore, do not aim to have clear conversion results like Fig.~\ref{fig:conversion-task}. \par 
% 7:27 SU
%
The second motivation lies in its application of font generation. We can find many past trials of automatic ``handwriting-style fonts'' designs, even before the deep-learning era. Our task can also be applied to the generation of handwriting-style fonts. As we will see later, our domain conversion method ``generates'' images that appear machine-printed from handwritten images. In other words, our method does ``not choose'' the best one from the existing font images for a given handwritten image.\par % 
The third motivation lies in the potential for developing a new OCR paradigm. Instead of recognizing handwritten characters directly, it will result in better accuracy by pre-transforming handwritten characters into their ``easier-to-read'' printed version. Moreover, when the main purpose of handwritten character recognition is the ``beautification of characters,'' just performing the conversion alone would fulfill the purpose. \par % 2/2 15:38 SU
The last motivation lies in a more fundamental question to understand what ``character classes'' are. As mentioned earlier, there are substantial shape differences between handwritten and machine-printed characters. However, convolutional neural networks (CNN) trained with a mixture of handwritten and printed character images can recognize characters in both domains without any degradation from the mixture~\cite{ide2017does,uchida2016further}. We humans also seem not to make a conscious differentiation between the two domains. The reason why we can read characters without differentiation remains, to the best of our knowledge, not fully elucidated. If the conversion is possible, it means that there is a mapping (correspondence) between handwritten and printed characters. This, in turn, may serve as one hypothesis to explain the ability to read the two domains without any differentiation.
\par  % 7:35 OK 

In machine learning-based image conversion, supervised learning is a common approach where
each training image is paired with its corresponding ideal transformed images before training. For instance, pix2pix~\cite{isola2017pix2pix} is a Generative Adversarial Network (GAN)~\cite{goodfellow2014gan}, where real images (such as photographic images) are paired with their semantic segmentation maps. By learning the inverse of regular segmentation processes, that is, by learning the conversion from the segmentation maps to the real images, it becomes possible to generate realistic images from segmentation maps. Various supervised image conversion approaches, including the more conventional encoder-decoder model like U-Net~\cite{unet}, have made it feasible to achieve image conversions that are highly challenging by traditional image processes.
\par %7:39 SU

CycleGAN~\cite{zhu2020cyclegan} and its variants make GAN-based image conversion much easier because 
they are free from the difficulty of preparing paired images. In fact, for our conversion task, the preparation of appropriate image pairs between two domains is not straightforward because there is no clear ground-truth. Since CycleGAN can learn the relationships between domains without explicit image correspondences, it is helpful for our conversion task. Murez~\textit{et al.}~\cite{murez2018image} already achieved domain conversion between scene-text images (Street View House Number, SVHN) and handwritten character images (MNIST) using CycleGAN. 
\par %  8:09 SU
\begin{figure}[t]
    \centering
    \includegraphics[width=\linewidth]{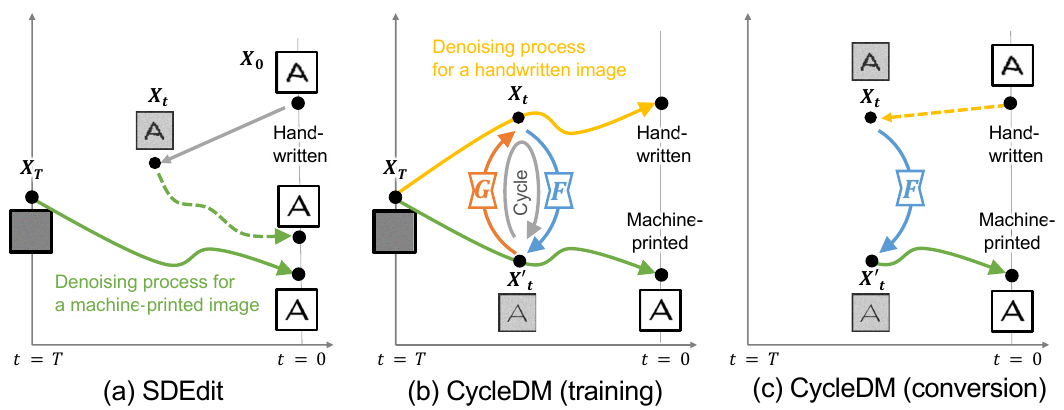}\\[-3mm]
    \caption{(a)~SDEdit~\cite{meng2022sdedit} for cross-domain conversion. Here, a handwritten character image is converted to its machine-printed version, but it is straightforward to realize the conversion in the reverse direction. (b) and (c)~Overview of the proposed CycleDM in its training phase and conversion phase, respectively.For simpler notations, $F_t$ and $G_t$ are used instead of $F_t$ and $G_t$.}
    \label{fig:overview}
\end{figure}

Diffusion models~\cite{dhariwal2021diffusion}, or DDPM~\cite{ho2020denoising}, are gaining attention as models capable of generating higher-quality images than GANs. A diffusion model uses an iterative denoising process starting from a purely-noise image. The main body of the model is a U-Net whose input is a noisy image and output is a noise component in the input image. By subtracting the estimated noise component from the input image, a less noisy image is obtained. Diffusion models have been applied to character image generation~\cite{he2023wordart,shirakawa2023,Tanveer_2023_ICCV}. Recently, image-conditioned diffusion models have also been realized~\cite{palette,wang2022semantic,Zhang_2023_ICCV}, making them applicable to image conversion as well.\par  % 2/2 19:50 SU
Then, a natural question arises --- {\em How can we realize correspondence-free image conversion with a diffusion model?} A simple answer to this question is to use SDEdit~\cite{meng2022sdedit}. As illustrated in Fig.~\ref{fig:overview}(a), SDEdit is a domain conversion technique that uses a pretrained diffusion model for a target domain. SDEdit assumes a noisy image of the source domain as that of the target domain and then starts the denoising process in the target domain. Therefore, the domain conversion becomes difficult if this assumption is not satisfied well. \par % 2/4 17:40 SU
In this paper, we propose a novel correspondence-free image conversion model called {\em CycleDM}, which utilizes the concept of CycleGAN in the diffusion model. More specifically, as shown in Fig.~\ref{fig:overview}(b), CycleDM uses not only pretrained diffusion models for both domains
but also two additional conversion models, $F_t$ and $G_t$, where $t$ is a specific iteration step of the denoising proces. These additional models allow conversion between two domains at $t$. As shown in Fig.~\ref{fig:overview}(c), after this explicit conversion, CycleDM can start its denoising process from $t$ in the target domain, more smoothly than SDEdit. \par
 % 8:07
 %
In the experiments, we evaluate the accuracy of cross-conversion between EMNIST, a handwritten character dataset, and Google Fonts, a machine-printed character image dataset. Through both qualitative and quantitative evaluations, we demonstrate that CycleDM can show better conversion quality than SDEdit, as well as CycleGAN. 
\par % 8:09 SU

The main contributions of this paper are summarized as follows:
\begin{itemize}
\item We develop a novel domain conversion model called CycleDM, which generates high-quality converted images based on diffusion models. 
\item While CycleDM can be applied to arbitrary image conversion tasks, we use it for the conversion task between two character image domains, that is, machine-printed and handwritten, according to the above multiple motivations.
\item Through both quantitative and qualitative evaluations, we confirmed that CycleDM enables cross-domain conversion far more accurately than SDEdit and CycleGAN. 
\end{itemize}
% 8:14 SU

% ===================================================================
\section{Related work} 
% ===================================================================
\subsection{Brief review of diffusion model}
The Diffusion Denoising Probabilistic Model (DDPM)~\cite{ho2020denoising} is a generative model that learns the process of transforming a purely-noise image into realistic images. DDPM consists of the diffusion process, where noise is progressively added to an image, and the denoising process, where noise is gradually removed.
\par % 8:16 SU
% The Diffusion Denoising Probabilistic Model (DDPM)~\cite{ho2020denoising} is a generative model that learns the process of transforming noise into images. DDPM consists of a \textit{forward process}, in which noise is progressively added to an image, and a \textit{reverse process}, in which noise is gradually removed, as illustrated in Figure~\ref{f:model_ddpm}.

In the diffusion process, noise is incrementally added to an image $X_0$ until it becomes a completely noisy image $X_T$ that follows a standard normal distribution $\mathcal{N}(0,1)$. The image $X_t$ at any point during the noise addition can be expressed as:
\begin{equation}
   X_{t} = \sqrt{\bar{\alpha}_{t}}X_0 + \sqrt{1 - \bar{\alpha}_{t}}\epsilon,\label{eq:diffusion}
\end{equation}
where $X_t$ denotes the noisy image after $t$ iterations of noise addition to the original image $X_0$. Here, $\epsilon$ represents random noise from a standard normal distribution $\mathcal{N}(0,1)$, and $\bar{\alpha}_{t}$ is calculated using a variance scheduler $\beta_t$ as $\bar{\alpha}_{t} = \prod_{s=1}^t (1-\beta_s)$, which controls the intensity of noise at each step $t$. (In the following experiment, the variance scheduler $\beta_t$ starts at $\beta_1=10^{-4}$ and linearly increases to $\beta_T=0.02$ over time.)
\par % 8:17 SU

During the denoising process, the purely-noise image $X_T$ is progressively transformed back into a clean image by gradually removing its noise component using a neural network model $\epsilon_{\theta}$, where $\theta$ represents the weight parameters of the model. Given $X_t$ and $t$, the output of the trained model ${\epsilon}_{\theta}(X_t, t)$ estimates the noise component $\epsilon$ added to $X_{t-1}$. The denoised image $X_{t-1}$ from one timestep earlier can be recovered using the following equation:
\begin{equation}
    X_{t-1} = \frac{1}{\sqrt{\alpha_t}}\left(X_t - \frac{1-\alpha_t}{\sqrt{1-\bar{\alpha}_t}}\epsilon_\theta(X_t,t)\right)+\sigma_t{z}, 
\end{equation}
where $\alpha_t = 1-\beta_t$, $\sigma_t = \sqrt{\beta_t}$, and $\bm{z}$ is additional random noise sampled from a standard normal distribution $\mathcal{N}(0,1)$. The denoised image $X_{t-1}$ is still a ``noisy'' image. However, at $t=0$, $X_0$ finally becomes a noiseless image. 
\par % 8:19 SU

To generate images, the model $\epsilon_\theta$ needs to accurately estimate the noise $\epsilon$ added to $X_t$. Furthermore, for conditional image generation, such as when specifying a particular class $c$ (e.g., style or character class), the model uses the conditional output $\epsilon_{\theta}(X_t, c, t)$ learned during training. Therefore, the model is trained to minimize the following loss function for conditional image generation:
\begin{equation}
    \mathcal{L}_{\mathrm{DDPM}} = \mathbb{E}_{X_0,c,\epsilon,t}\left[\|\epsilon-\epsilon_{\theta}(X_t,c,t)\|_2^2\right].
\end{equation}
Here, $c$ denotes a specific class associated with the original image $X_0$. The noise $\epsilon$ is sampled from a standard normal distribution $\mathcal{N}(0,1)$, and $t$ is sampled from a uniform distribution $\mathcal{U}(1,T)$. The noisy image $X_t$ is derived from $X_0$, $\epsilon$, and $t$ according to the described noise addition process.

% -------------------------------------------------------------------
\subsection{Diffusion models for image conversion}
% 1. Diffusion modelでのimage conversionとして考えられるのは，SDEdit, Image-to-image modelに特化させたモデルでの変換である．
% SDEditでの変換は～である．追加の学習を必要とせずある程度形状を捉えた変換が可能である．
% 2. 次に考えられるのは，Image-to-imageに特化させたモデルである．Cyclediffusionでは．これらの殆どは，textureによる変換がほとんどで文字変換に対応していないと考えられる．
% 3.  本研究では，CycleGANとdiffusion modelを融合させるという今までにないアプローチで文字変換を可能にしていた新たなモデルを提案している．
%SDEdit and more
% CycleDiffusion
% StainDiff: Transfer Stain Styles
Diffusion models for image conversion are mainly divided into SDEdit and image-to-image translation models. First, SDEdit~\cite{meng2022sdedit} is a well-known usage of diffusion models for image conversion. The overview of this method is already shown in Fig.~\ref{fig:overview}~(a). It involves adding a specific level of noise to a certain image $X_0$ in the source domain to ``mimic'' a noisy image $X'_t$ of the target domain. Then $X'_t$ is denoised through the denoising process of the target domain. Finally, $X'_0$ is given as the final result in the target domain. SDEdit is a powerful domain conversion technique in the sense that it does not require any additional training. However, shown in Fig.~\ref{fig:overview}~(a), the noisy image $X'_t$, from which the denoising process starts, is not a real noisy image in the denoising process of the target domain and deviates from the real noisy images. This deviation often causes unrealistic $X'_0$, as we will see in our experiment.\par
% SU 8:14

Second, image-to-image translation models focus on image conversion, unlike SDEdit. Sasaki~\textit{et al.}~\cite{sasaki2021unit} and Wu~\textit{et al.}~\cite{Wu_2023_ICCV} and proposed diffusion models that leverage latent space for image conversion. Moreover, diffusion models have been widely applied for style transfer, which is one of the image conversion tasks. Shen~\textit{et al.}~\cite{staindiff} translated stain styles of histology images by diffusion models. A lot of studies~\cite{ahn2023dreamstyler,controlstyle,Hamazaspyan_2023_CVPR,DiffStyler,Pan_2023_WACV,SGDiff,yang2023zero} generated synthetic images that are faithful to the style of the reference image in addition to the prompt in the text-to-image model (e.g., Stable Diffusion~\cite{Rombach_2022_CVPR}). These models aim for photographic image conversion and not for cyclic shape conversion.
\par

% 悪知恵
% このように数多くのDiffusion modelによるimage conversionに特化した手法が提案されている．しかしながら，これらの手法は，文字変換(=形状や局所形状)を想定したモデルではないため，比較手法として適していないと考えられる．そこで，本研究では，形状変換にも対応しているSDEditのみをDiffusion modelによるimage conversion手法の比較対象とする．
% 
% SDEdit
% 正確には，今回の手法はCycleGANとSDEditを組み合わせることで，それぞれの手法よりも高精度に画像変換を行うことを目的とする
% -------------------------------------------------------------------
\vspace{-2mm}
\subsection{Diffusion models for character image generation}
Diffusion models have also been applied to the various character image generation tasks. 
Gui~\textit{et al.}~\cite{gui2023zero} tackled a zero-shot handwritten Chinese character image generation with DDPM for creating training data for OCR. For machine-printed character image generation, especially few-shot font generation, Yang~\textit{et al.}~\cite{yang2024fontdiffuser,he2022diff} and He~\textit{et al.}~\cite{he2022diff} leveraged diffusion models. In addition to typical handwriting characters and machine-printed characters, diffusion models have also been used for artistic typography generation~\cite{wang2023anything,Tanveer_2023_ICCV,IluzVinker2023,shirakawa2023}.
\par % 8:47 SU

Our CycleDM is a novel model that uses the concept of CycleGAN in the diffusion models for unpaired image conversion, where no image correspondence is necessary between two domains.
Moreover, no diffusion model has been applied to cross-domain conversion between machine-printed character images.
\par % 8:45 SU

% ===================================================================
\section{CycleDM}
% ===================================================================
% pptx/cycleDM-loss.pptx
\begin{figure}[t]
    \centering
    \includegraphics[width=1.0\linewidth]{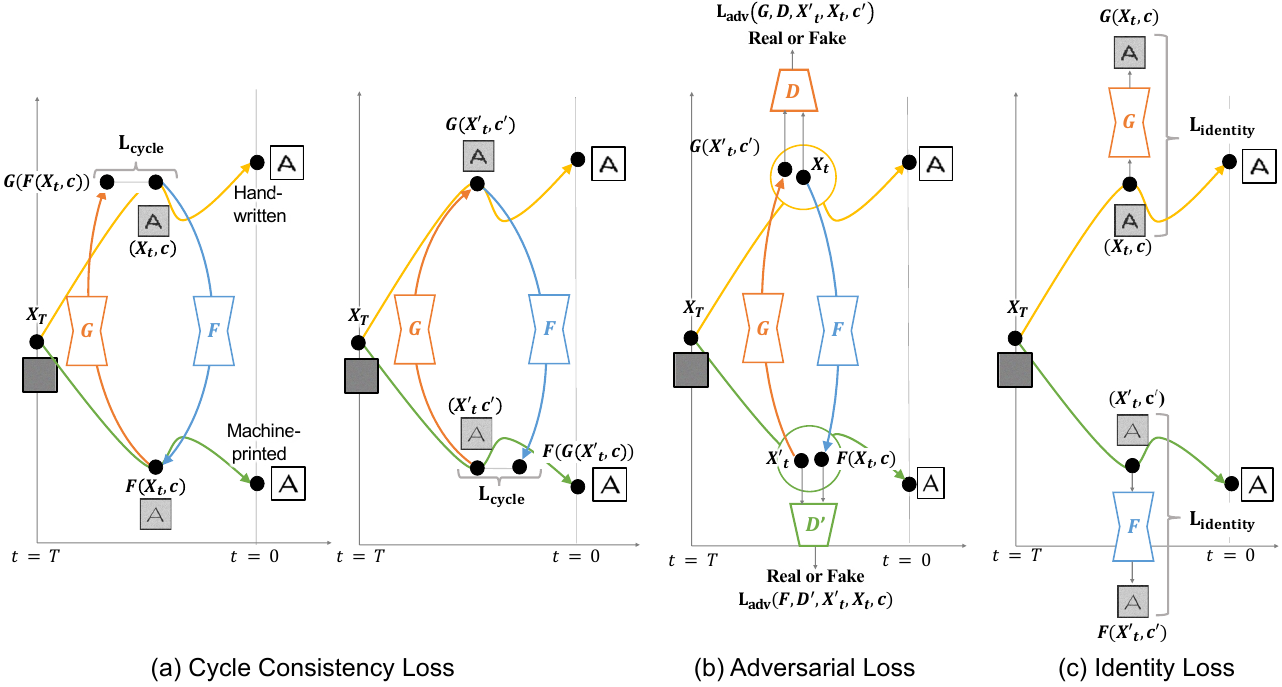}\\[-3mm]
    \caption{Loss functions to train the conversion models $F_t$ and $G_t$ of CycleDM. 
    For simplicity, $F_t$ and $G_t$ are denoted as $F$ and $G$ and the class condition $c$ is omitted. The backbone DDPM is pretrained, and its parameters are frozen during training  $F_t$ and $G_t$.}
    \label{fig:model_loss}
\end{figure}
% -------------------------------------------------------------------
\subsection{Overview\label{sec:overview}}
In this section, we detail CycleDM, a new method for unpaired cross-domain image-to-image conversion. CycleDM introduces the concept of CycleGAN~\cite{zhu2020cyclegan} into DDPM~\cite{ho2020denoising} for realizing higher-quality image conversions without any correspondence between the two domains.
\par 
%8:52 SU

Fig.~\ref{fig:overview}(b) shows the principle of CycleDM. Assume that
we have a conditional DDPM pretrained to generate images of both domains. The conditions of DDPM are the domain, the class $c$, and the time index $t$ of the denoising process. For the character image conversion task of Fig.~\ref{fig:conversion-task}, each domain is machine-printed or handwritten, $c\in \{`A,'\ldots, `Z'\}$, and $t\in \{0,\ldots, T\}.$  Hereafter, $X_t$ denotes a noisy image under the denoising process at $t$ in the handwritten character domain, whereas $X'_t$ is a noisy image in the machine-printed character domain. 
Consequently, $X_0$ and $X'_0$ denote final (i.e., completely denoised) images.
When the domain condition is ``handwritten,'' the DDPM will generate a handwritten image $X_0$ from a purely-noise image $X_T$ via $X_t$. When the condition is ``machine-printed,'' it generates a machine-printed image $X'_0$ from $X_T$ via $X'_t$. 
Once the DDPM is trained to generate handwritten and machine-printed character images, its model parameters are frozen during the later process (to train $F_t$ and $G_t$). \par
% 8:57 SU
%
The unique mechanism of CycleDM is two conversion models, $F_t$ and $G_t$, each of which is a model in a convolutional encoder-decoder structure conditioned by a character class $c$.
Specifically, the model $F_t(X_t,c)$ converts $X_t$ into $X'_t$, whereas $G_t(X'_t,c)$ converts $X'_t$ into $X_t$. Consequently, these conversion models realize the interchangeability between two domains. As detailed in Section~\ref{sec:training}, these conversion models are trained under the pretrained DDPM, without any image-to-image correspondence between $X_t$ and $X'_t$. 
Note that the conversion models are denoted as $F_t$ and $G_t$ rather than $F$ and $G$; this is because they need to be trained for the conversion at a certain step $t$. 
\par
% 9:05 SU
%
Fig.~\ref{fig:overview}(c) shows the process of the cross-modal conversion using the trained CycleDM. Assume a task to convert a handwritten image $X_0$ to its (unknown) machine-printed version. In this task, $X_0$ is {\em diffused} to be a noisy image $X_t$, like SDEdit. This diffusion process follows Eq.~(\ref{eq:diffusion}). Then, by using $F_t$, a converted version of $X_t$ is given as $X'_t = F_t(X_t, c)$. Finally, the machine-printed version is generated by the denoising process from $X'_t$ to $X'_0$ by the DDPM in the machine-printed character domain
with the condition $c$. The conversion from a machine-printed character image $X'_0$ to its (unknown) handwritten version is also possible by the denoising process.
\par % 9:07 SU

%
% -------------------------------------------------------------------
\subsection{Training CycleDM\label{sec:training}}
As noted before, we do not know the ground-truth of the conversion pair $X_0$ and $X'_0$. In other words, we do not know which machine-printed image $X'_0$ is appropriate as the conversion result of a handwritten image $X_0$, and vice versa. We, therefore, need to train $F_t$ and $G_t$ without giving ideal conversion pairs of $X_0$ and $X'_0$. Fortunately, we can employ the concept called cycle consistency, which is used in CycleGAN, for training $F_t$ and $G_t$ without any pairs. \par
% 9:08 SU
%
\subsubsection{Cycle Consistency Loss}  The cycle consistency is defined as the relations $F_t(G_t(X'_t, c')) \approx X'_t$ and $G_t(F_t(X_t,c)) \approx X_t$ for any $X_t$ and $X'_t$. Each relation depends only on $X_t$ or $X'_t$; therefore, we do not need to use any correspondence between $X_t$ and $X'_t$ to evaluate how the relations are satisfied. More specifically, as shown in Fig.~\ref{fig:model_loss}(a), we introduce the cycle consistency loss $\mathcal{L}_{\mathrm{cycle}}$, which evaluates how the cycle consistency is unsatisfied:
\begin{eqnarray}
    \mathcal{L}_{\mathrm{cycle}}(F_t,G_t,c,c') &=&\mathbb{E}_{X_t,c}\left[ \| G_t(F_t(X_t,c)) - X_t \|_1 \right]\nonumber \\
    && +\mathbb{E}_{X'_t,c'}\left[ \| F_t(G_t(X'_t,c')) - X'_t \|_1 \right].
\end{eqnarray}
% 9:12 OK SU

\subsubsection{Adversarial Loss} The conversion models $F_t$ and $G_t$ need to output realistic noisy images, $X'_t$ and $X_t$, respectively, at individual domains. This need is incorporated by the adversarial loss, which is a typical loss function of GANs. The adversarial loss uses two domain discriminators, $D$ and $D'$, each of which is a CNN to be trained with $F_t$ and $G_t$.
As shown in Fig.~\ref{fig:model_loss}(b), the discriminator $D$ needs to discriminate whether its input is $X_t$ (a real noisy image of the domain) or $G_t(X'_t,c')$ (a fake noisy image converted from the other domain), whereas $D'$ discriminates $X'_t$ and $F_t(X_t, c)$. 
Formally, the adversarial loss $L_{\mathrm{adv}}$ for the model $G_t$ and $D$ is defined as follows:
\begin{eqnarray}
\mathcal{L}_{\mathrm{adv}}(G_t, D, X'_t, X_t, c') 
  &=& \mathbb{E}_{X_t,c'}[\log D(X_t,c')] \nonumber\\
&&+ \mathbb{E}_{X'_t,c'}[\log(1 - D(G_t(X'_t,c')))].
\end{eqnarray}
\noindent Similarly, we have the loss for $F_t$ and $D'$ as 
$\mathcal{L}_{\mathrm{adv}}(F_t, D', X_t, X'_t, c)$. 
Minimizing these loss functions enables $F_t$ and $G_t$ to generate noisy images resembling those of their respective output domains.
\par
% 9:14 OK SU

\subsubsection{Identity Loss} For high-quality image conversion, $F_t$ and $G_t$ need not make any unnecessary changes. If their input already appears to belong to the target domain, they need to do anything. This means they must behave as an identical mapping if a noisy image in the target domain is input. More formally, as shown in Fig.~\ref{fig:model_loss}(c), $F_t$ and $G_t$ need to satisfy $F_t(X'_t, c)\approx X'_t$ and $G_t(X_t, c)\approx X_t$. For this purpose, the following identity loss is introduced:
\begin{eqnarray}
\mathcal{L}_{\mathrm{identity}}(F_t,G_t,c,c') &=&\mathbb{E}_{X'_t, c'}\left[ \| F_t(X'_t,c') - X'_t \|_1 \right]\nonumber \\
    &&+\mathbb{E}_{X_t,c,}\left[ \| G_t(X_t,c) - X_t \|_1 \right]
\end{eqnarray}
% 9:16 OK SU

\subsubsection{Training the conversion models $F_t$ and $G_t$}
The final loss function $\mathcal{L}_{\mathrm{total}}$ for training $F_t$, $G_t$ (as well as $D$ and $D'$) is given by:
\begin{eqnarray}
     \mathcal{L}_{\mathrm{total}}(F_t,G_t,D,D',c,c') &=& \mathcal{L}_{\mathrm{adv}}(F_t, D', X_t, X'_t,c) \nonumber\\
     &&+ \mathcal{L}_{\mathrm{adv}}(G_t, D, X'_t, X_t,c')\nonumber \\
     &&+ \lambda_{\mathrm{cycle}} \mathcal{L}_{\mathrm{cycle}}(F_t,G_t,c,c')  \nonumber \\
     &&+ \lambda_{\mathrm{identity}} \mathcal{L}_{\mathrm{identity}}(F_t,G_t,c,c'), 
\end{eqnarray}
where $\lambda_{\mathrm{cycle}}$ and $\lambda_{\mathrm{identity}}$ are weight coefficients. As noted before, the conditional DDPM is pretrained to produce $X_t$ and $X'_t$, and then its model parameters are frozen (i.e., not updated) during training $F_t$ and $G_t$. To stabilize the training, Gradient Penalty~\cite{ishaan2017wgan} is applied to the training of the discriminators $D$ and $D'$. 
% OK 9:19 SU

% ===================================================================
\section{Experimental Results}
% ===================================================================
Hereafter, we abbreviate handwritten character images and machine-printed character images as HWs and MPs, respectively, for simplicity. 

\subsection{Dataset}
% -------------------------------------------------------------------
We use the EMNIST dataset~\cite{cohen2017emnist} for HWs and the Google Fonts dataset \footnote{\url{https://github.com/google/fonts}} for MPs. This paper assumes characters from 26 classes of the capital Latin alphabet, although CycleDM can deal with other alphabets (and even more general images, like cat and dog images). The EMNIST dataset comprises {27,600} capital letter images (about {1,062} for each of 26 classes), and the Google Fonts dataset comprises {67,600} capital letter images ({2,600} different fonts for each letter class). 
\par
% 9:35 SU
%
The datasets are split for 70\% training and 30\% testing.
Specifically, the EMNIST dataset is split for training ({19,308} images) and testing ({8,192}), whereas the Google Fonts dataset for 
training ({47,294}) and testing ({20,306}). The training images are used for training not only DDPM but also the conversion models $F_t$ and $G_t$. 
As the conversion test, the {8,192} test images from the EMNIST are converted to their MP version by CycleDM, and similarly, the {20,306} test images from the Google Fonts are converted to their HW version.
\par
% 9:35 SU

%The DDPM used in this method was trained with 19,308 handwritten character images and 47,294 Machine-printed character images.

% 生成された文字画像を評価するために，テストデータとして手書きから活字への変換ではEMNISTの大文字26クラスから8192枚の手書き文字画像を使用し，活字から手書き文字への変換ではGoogle Fontsの大文字26クラスから20224枚の活字文字画像を使用した．

% -------------------------------------------------------------------
\subsection{Implementation details}
% -------------------------------------------------------------------
The diffusion model was trained with total steps $T=1000$, a batch size of 64, across 200 epochs. The conversion models $F_t$ and $G_t$ follow the encoder-decoder structure used in CycleGAN\cite{zhu2020cyclegan}. These conversion models are trained over 100 epochs with a batch size of 64. The weights to balance the loss functions were set at $\lambda_\mathrm{cycle}=2.0$ and $\lambda_\mathrm{identity}=1.0$ by a preliminary experiment.\par
% 9:45 SU
%
The step $t$ for the conversion modules $F_t$ and $G_t$ is important for our work. If $t$ is close to $T (=1,000)$, the conversion will be made on more noisy images, and therefore, the difference between the source and result images can be large. (This is because the appearance of the source image will mostly disappear by adding a large amount of noise.) In contrast, if $t$ is close to $0$, the difference will be small. In the following experiment, we prepare the conversion modules at $t=400, 500$, and $600$, following the suggestion in SDEdit~\cite{meng2022sdedit}.
% 9:50 SU
%% -------------------------------------------------------------------
\subsection{Comparative methods}
% -------------------------------------------------------------------
% 
%比較手法として，拡散モデルを使用した画像変換の手法であるSDEdit\cite{meng2022sdedit}とCycleGAN\cite{zhu2020cyclegan}を採用した．
%SDEditは，提案手法と同条件で画像変換を行うために，文字クラスを条件として学習した条件付きDDPMを用いて，画像変換を行った．
%CycleGANも同様に，文字クラスのClass-Conditionaを与えて学習を行ったモデルを採用した．
%Chenlinらは，SDEditは画像変換においてt=300~600が望ましい時刻tであると唱えており，我々もこれに従い，時刻t=400~600で生成を行った

We used the following two conversion models as comparative methods. 

% \subsubsection{Baseline}
% One naive idea of domain conversion is the nearest-neighbor search.
% For example, if a handwritten image of `A' is given, we assume that the nearest machine-printed image in all the machine-printed images in the training set is its converted result. This does not generate any new images in the target domain, and therefore, it 
% has an inherently different purpose from domain conversion tasks. 
% For the nearest-neighbor search, we use the $L_1$ distance between two (bitmap) character images.

\subsubsection{SDEdit}
In the experimental setup,  SDEdit differs from our CycleDM only in the absence of the conversion module before the denoising process in the target domain. (Recall Fig.~\ref{fig:overview} (a) and (c).) The other setup is the same; namely, the same DDPM is used for denoising. 
For the SDEdit, we also examine three different points $t=400, 500$, and $600$ for starting the denoising process.\par
% 10:00 ぐらい？ SU

% Cycleconsistencyを有する比較手法としてCycleleGANを採用した
% 本実験では，通常のCycleGANとは異なり，文字クラスに関してconditional GANのような条件付けを行った．
% また，Generator, Discriminatorのアーキテクチャは，CyclGANのモデルに対して表現能力の改善のためにresidual layerを使用している．
% さらに，学習データは提案方法と同じである．
\subsubsection{CycleGAN}
CycleGAN is selected as a comparative method because it has domain conversion models $F$ and $G$ like our CycleDM. The backbone GAN discriminator and generator are modified from the original architecture to incorporate residual layers for better generation ability. The architectures of the conversion models are the same as CycleDM. For a fair comparison, we also introduce the class condition $c$ to its modules. 
This CycleGAN is trained by using the same training sets as CycleDM.
\par
% 12:23 SU (仮)
%-------------------------------------------------------------------
\subsection{Evaluation metrics\label{sec:metrics}}
%生成画像の定量評価には，手書き文字と活字を学習したcnnで特徴抽出を行い精度を比較する．
%具体的には，手書きから活字変換においては，テストデータである手書き画像から変換された画像と、テストデータの活字画像から得られる特徴量を比較し，生成画像のPrecison，Recall，FIDを算出した．

%また，定量評価では変換された画像と本来の文字画像との最近傍法による認識率の結果も提示し，生成された画像がどの程度正確に文字の形状を再現できているか評価した．

We employed FID~\cite{FID}, precision, and recall~\cite{kynkaanniemi2019improved} for quantitative evaluation.
FID is a standard metric that measures the distance between generated images and real ones in feature space for evaluating diversity and fidelity. Precision and recall also evaluate fidelity and diversity, respectively, using the feature space by EfficientNet, which is trained for classifying HWs and MPs. Roughly speaking, this precision and recall~\cite{kynkaanniemi2019improved} measure the overlap between the original and generated image distributions in the feature space. If both distributions are identical, precision and recall become one.
\par
% 10:39 SU

To evaluate the readability of generated images, we measure the accuracy of classifying the character class. This evaluation used the nearest-neighbor search in pixel with $L_1$ distance. The images to be searched are the test character images in the target domain. \par
% 10:44 SU
% -------------------------------------------------------------------
\subsection{Quantitative evaluations}
% -------------------------------------------------------------------

\begin{table}[t]
\centering
\caption{Quantitative evaluation of conversion from the handwritten character domain to the machine-printed character domain (HW$\to$MP).}\vspace{-2mm}
\label{g:mp_result}
\begin{tabular}{rcccc}
\hline
\multicolumn{1}{c}{Method} & Accuracy\(\uparrow\) & Precision\(\uparrow\) & Recall\(\uparrow\) & FID\(\downarrow\) \\ \hline
CycleGAN w/ Class-Condition & \underline{0.95} & \textbf{0.90} & 0.47 & 1.07 \\ 
Class-Conditonal SDEdit (t=400) & 0.53 & 0.57 & 0.80 & 0.99 \\
 (t=500) & 0.67 & 0.69 & 0.83 & 0.45 \\
 (t=600) & 0.88 & 0.82 & 0.85 & 0.23 \\ \hline
CycleDM (t=400) & 0.91 & 0.86 & \textbf{0.87} & 0.20 \\
(t=500) & 0.87 & \underline{0.87} & \underline{0.86} & \textbf{0.15} \\
(t=600) & \textbf{0.96} & \textbf{0.90} & 0.85 & \underline{0.16} \\ \hline
\end{tabular}
\end{table}
%表\ref{g:mp_result}に手書き画像から変換された活字画像の定量評価を示す．
% この表から，どの評価基準においても提案手法がtに依存せずに比較手法と同程度あるいはそれ以上のパフォーマンスを示している．
% 特に，提案手法ではどの時刻tにおいてもprecisionとrecallのバランスが取れている．これは，提案法がバリエーション豊富かつ活字らしい文字を生成できていることを示している．
% また，活字画像との最近傍法による認識率の値も，提案法が最も良い値になっており，手書き文字の形状を正確に捉えた活字文字が生成されているといえる．
% 一方，比較手法であるCycleGANはprecisionにおいて最も良い値を示している．ただし，はRecallの値が，最も低くなっている．
% これは，活字らしい形状の文字は生成できているものの，元の手書き画像とは異なる活字画像が生成されていることにを意味している．
% SDEditもRecallの値は高いことから様々なスタイルの活字を生成しているといえる．ただし，提案法に比べてPrecisionの値が劣っており，活字らしい画像の生成は提案法に劣る．
% SDEditはaccuracyやprecisionがtが大きくなると劇的に良くなる．
% 提案法はtが変わってもそこまで変わらないので，SDEditよりは安定して生成できる．
In this section, we show several quantitative evaluation results. They are rather macroscopic evaluations to observe how the ``set'' of generated images is appropriate in the target domain. Therefore, we used the metrics in Section~\ref{sec:metrics} for the macroscopic evaluations. A more microscopic evaluation to see how the converted images in the target domain hold their original images in the source domain will be made qualitatively in the next section. 
\par
% 12:12 SU

\subsubsection{Conversion from HW to MP}
Table~\ref{g:mp_result} shows the result of the quantitative evaluation of converting HW to MP.  From this table, it is evident that CycleDM has the best or near-best performance with others in all $t$ and all metrics. In particular, CycleDM has a good balance between precision and recall in all $t$.
This indicates that the generated MPs by CycleDM have not only similar appearances to real MPs but also diverse appearances that cover the real MPs.
Furthermore, the nearest neighbor recognition result shows that CycleDM achieves the best or near-best accuracy to the other methods regardless of the $t$.
This also indicates that MPs converted from HWs accurately mimic the style of the real MPs.
% 10:58 SU

% FID value, which measures the closeness of the distribution between the generated and original machine-printed images, and the Recall, which measures the coverage of various machine-printed images, are better than those of the comparison methods. Therefore, the machine-printed images generated by the proposed method have the most machine-printed-like character shapes and generate a variety of font styles.

Although CycleGAN shows the best performance in precision, it should be noted that CycleGAN has the lowest recall. This indicates that CycleGAN often generates MPs with similar appearances; in other words, generated MPs have less diversity than the real MPs. One might suppose this is reasonable because the diversity of HWs is not as large as that of MPs, and thus, the distribution of the generated MPs must be smaller than that of the real MPs. However, this is not correct; the later qualitative evaluations show that generated MPs by CycleGAN do not even reflect the original appearance of given HPs and show rather standard MP styles only.
\par
% 11:09 SU

SDEdit is the opposite of CycleGAN; SDEdit shows a high recall but a low precision. This indicates that SDEdit generates MPs with various styles but sometimes generates unrealistic MPs. The later qualitative evaluation will also confirm this observation. 
Note that SDEdit shows largely different performance by $t$, whereas CycleDM does not. This indicates that CycleDM is more stable than SDEdit.
\par
% 11:18 SU
% 手書き画像(test)と手書き画像(train)の最近傍法による認識率：0.87
%CycleDM(t=400)と活字画像(train)の最近傍法による認識率：0.92
%CycleDM(t=500)と活字画像(train)の最近傍法による認識率：0.90
%CycleDM(t=600)と活字画像(train)の最近傍法による認識率：0.97
%SDEdit(t=400)と活字画像(train)の最近傍法による認識率：0.55
%SDEdit(t=500)と活字画像(train)の最近傍法による認識率：0.70
%SDEdit(t=600)と活字画像(train)の最近傍法による認識率：0.90
%CycleGANと活字画像(train)の最近傍法による認識率：0.96
\subsubsection{Does conversion from HW to MP help OCR?}
As noted in Section~\ref{sec:intro}, one of our motivations is that conversion from HW to MP will be a good preprocessing for OCR. To confirm the positive effect of the conversion, we conducted 
26-class character classification experiments. As the classifier, we use a simple but intuitive nearest-neighbor search with $L_1$ distance. When we classify the original test HWs with the original training HWs, the classification accuracy was about 87\%. In contrast, when we convert HWs to MPs and then classify them with the original training MPs, the accuracy rises up to about 97\%. This result simply suggests the conversion helps OCR.

% \begin{table}[t]
% \centering
% \caption{活字生成後の活字クラスごとの生成評価}
% \label{g:mp_class_result}
% \begin{tabular}{lccc}
% \hline
% Method & Precision\(\uparrow\) & Recall\(\uparrow\) & FID\(\downarrow\) \\ \hline
% Class-Conditonal SDEdit (t=400) & 0.56 & 0.77 & 3.46 \\
% Class-Conditonal SDEdit (t=500) & 0.66 & 0.84 & 1.86 \\
% Class-Conditional SDEdit (t=600) & 0.81 & 0.86 & 0.68 \\
% CycleGAN w/ Class-Condition & 0.88 & 0.55 & 1.24 \\ \hline
% Ours (t=400) & 0.84 & \textbf{0.89} & \textbf{0.39} \\
% Ours (t=500) & 0.85 & 0.87 & 0.55 \\
% Ours (t=600) & \textbf{0.90} & 0.86 & 0.42 \\ \hline
% \end{tabular}
% \end{table}

\begin{table}[t]
\centering
\caption{Quantitative evaluation of conversion from the machine-printed character domain to the handwritten character domain (MP$\to$HW).}\vspace{-2mm}
\label{g:hw_result}
\begin{tabular}{rcccc}
\hline
\multicolumn{1}{c}{Method}& Accuracy\(\uparrow\) & Precision\(\uparrow\) & Recall\(\uparrow\) & FID\(\downarrow\) \\ \hline
CycleGAN w/ Class-Condition & \underline{0.81} & \underline{0.87} & 0.81 & \textbf{0.10} \\ 
Class-Conditonal SDEdit (t=400) & 0.67 & 0.80 & 0.70 & 0.79 \\
 (t=500)  & 0.72 & 0.83 & 0.76 & 0.60 \\
 (t=600) & 0.79 & 0.84 & \underline{0.82} & 0.55 \\ \hline
CycleDM (t=400) & 0.80 & \textbf{0.88} & \underline{0.82} & \underline{0.11} \\
(t=500) & 0.79 & \underline{0.87} & \textbf{0.83} & \underline{0.11} \\
(t=600) & \textbf{0.85} & \textbf{0.88} & \textbf{0.83} & 0.13 \\ \hline
\end{tabular}
\end{table}

%表\ref{g:hw_result}に活字画像から変換された手書き画像の定量評価を示す．
%  表から提案法がどの時刻&どのmetricでも大体best
%  t=400の時はprecisionとFIDがbestなのに，t=600ではaccuracyがbest． これは，handwriting styleと読みやすさの関係んいトレードオフがある可能性あり (無理やりすぎ？)
%CycleGANもPrecision，Recall，FIDの値が良いことから，手書き文字らしい形状かつ様々なスタイルの手書き画像を生成している．しかし，表\ref{g:mp_result}において，Recallの値が極端に低いことから，本研究の目的である文字画像の相互変換においては提案法がより優れているといえる．

\subsubsection{Conversion from MP to HW}
Table \ref{g:hw_result} shows the results of the quantitative evaluation of HWs converted from MPs. Again, CycleDM achieves the best or near-best performance with others in all $t$ and all metrics. Comparing CycleDM to SDEdit, CycleDM is more stable to $t$ like the previous setup. High FID values of SDEdit show the difficulty of generating realistic HWs for SDEDit. On the other hand, different from the previous setup of HW$\to$MP, CycleGAN shows a good recall in this setup. This is because the diversity of real HWs is rather small than that of real MPs, and thus, CycleGAN could ``abuse'' its non-diverse generation ability for better recall. 
\par 
% 11:33 SU

% \begin{table}[t]
% \centering
% \caption{手書き文字生成後の手書きクラスごとの生成評価}
% \label{g:hw_class_result}
% \begin{tabular}{lccc}
% \hline
% Method & Precision\(\uparrow\) & Recall\(\uparrow\) & FID\(\downarrow\) \\ \hline
% Class-Conditonal SDEdit (t=400) & 0.80 & 0.69 & 2.77 \\
% Class-Conditonal SDEdit (t=500) & 0.83 & 0.79 & 2.00 \\
% Class-Conditional SDEdit (t=600) & 0.85 & 0.83 & 1.55 \\
% CycleGAN w/ Class-Condition & \textbf{0.90} & 0.79 & 0.59 \\ \hline
% Ours (t=400) & \textbf{0.90} & 0.82 & 0.54 \\
% Ours (t=500) & 0.89 & \textbf{0.84} & 0.58 \\
% Ours (t=600) & \textbf{0.90} & 0.83 & \textbf{0.53} \\ \hline
% \end{tabular}
% \end{table}

% \begin{table}[t]
% \centering
% \caption{unconditional modeにおける活字認識}
% \label{g:result}
% \begin{tabular}{lccc}
% \hline
% Method & Accuracy \(\uparrow\) & FID\(\downarrow\) \\ \hline
% %変換前の手書き画像がBaseline
% Baseline & 0.54 & 4.58  \\ 
% SDEdit (t=400) & 0.49 & 3.57  \\
% SDEdit (t=500) & 0.39 & 3.18 \\
% SDEdit (t=600) & 0.19 & 4.44 \\ \hline
% Ours (t=400)& \textbf{0.84} & \textbf{0.54} \\
% Ours (t=500) & 0.51 & 1.90 \\
% Ours (t=600) & 0.17 & 4.78 \\ \hline
% \end{tabular}
% \end{table}

\begin{table}[t]
\centering
\caption{Quantitative evaluation of the conversion from the handwritten character domain to the machine-printed character domain (HW$\to$MP) without the class condition $c$.  
% The Baseline shows handwritten characters before converting machine-printed characters. Precision, recall, and FID are not available in the Baseline because the Baseline follows real image distribution.
}\vspace{-2mm}
\label{g:uncond_result}
\begin{tabular}{rccccc}
\hline
\multicolumn{1}{c}{Method} & Accuracy \(\uparrow\) & Precision \(\uparrow\) & Recall \(\uparrow\) & FID\(\downarrow\) \\ \hline
% 変換前の手書き画像がBaseline
% Baseline & 0.54 & --  & -- &  -- \\ \hline
SDEdit (t=400) & 0.49 & 0.56 & 0.79 & 0.99  \\
 (t=500) & 0.39 & 0.67 & 0.84 & 0.51 \\
 (t=600) & 0.19 & 0.78 & \textbf{0.87} & 0.28 \\ \hline
CycleDM (t=400) & \textbf{0.84} & 0.86  &\textbf{0.87} & 0.18  \\
 (t=500) & \underline{0.51} & \underline{0.88} & \underline{0.85} & \underline{0.16}  \\
 (t=600) & 0.17 & \textbf{0.90} & \textbf{0.87}  & \textbf{0.10} \\ \hline
\end{tabular}
\end{table}
%unconditional mode
%この表は，文字クラスのClass-Conditionを与えて学習した変換モデル$F_t,G_t$及びDDPMを条件なしで，手書き画像から変換された活字画像の定量評価を示す．Baselineは変換前の手書き画像を示す．
% この表から，t=400において提案法の最近傍法による認識率の結果が最も高いことがわかる．このことから，t=400ではある程度の活字認識ができていることが分かる．
% ただし，t=500やt=600においてはBaselineである手書き画像と活字画像の最近傍法による認識率よりも精度が悪くなっていることが分かる．この原因として，付与されたノイズの大きさが考えられる．
% $t=400$に比べて，$t=500$や$t=600$はノイズの強い画像となっており，条件なしのDDPMでデノイズをした時に別な文字クラスの画像が生成されたことが考えられる．
% そのため，変換モデル$F,G$が時刻t時点ではある程度の手書き画像を認識しているが，条件なしDDPMによって認識率が下がったと考えられる．
%SDEditの最近傍法による認識率も時刻$t$が高くなるにつれて認識率の結果が悪くなっていることから，ノイズの追加による認識の影響は大きいといえる．

\subsubsection{Is the class condition $c$ important for conversion?}
In the above experiments, we always gave the class condition $c$
to all the models, i.e., $F_t$, $G_t$, and DDPM. For example, when we convert an MP `A' to its HW version, we input the condition $c=$`A' for all the models. (Of course, the comparative models, SDEdit and CycleGAN, also used the same class condition $c$ for a fair comparison). One might think, ``The good performance of CycleDM comes from the class condition $c$ -- So, without $c$, the performance might be degraded drastically.''
\par 
% 12:03 SU
Table \ref{g:uncond_result} proves that, at least for our CycleDM, the class condition $c$ is important but not by much. This table shows the result of the quantitative evaluation of the conversion from HWs to MPs without class condition $c$ in $F_t$, $G_t$, and DDPM.\footnote{In diffusion models, ignorance of a specific condition is realized by feeding a ``null token'' instead of a real condition. The null token is a near-random vector trained along with the model under their unconditional mode.} From this table, CycleDM still achieves the recognition accuracy 84\%; compared to the accuracy 96\% in Table~\ref{g:mp_result}, it is a large degradation, but still comparable to 88\% by SDEdit with the class condition. (The accuracy of SDEdit drops down to 49\% without the class condition.) This result suggests that the conversion models $F_t$ and $G_t$ naturally manage class differences in the noisy image space.
Moreover, it should be emphasized that CycleDM could keep its high recall and precision and low FID as Table~\ref{g:mp_result}, even without the class condition. 
\par
% 12:03 SU 

% This could be attributed to the amount of added noise. 
% Compared to $t=400$, images at $t=500$ and $t=600$ are noisier, and it is possible that images of different character classes were generated when denoising was performed with DDPM without condition.
% Therefore, it can be inferred that while the conversion models $F, G$ recognize a certain level of handwritten character images at time $t$, the recognition rate decreased due to denoising by the condition-less DDPM. 
% The recognition accuracy results of SDEdit also worsen as the time $t$ increases, indicating that the impact of added noise on recognition is significant.

% -------------------------------------------------------------------
\subsection{Qualitative evaluation\label{sec:qualitative}}
% -------------------------------------------------------------------
\begin{figure}[t]
    \centering
    \includegraphics[width=0.8\linewidth]{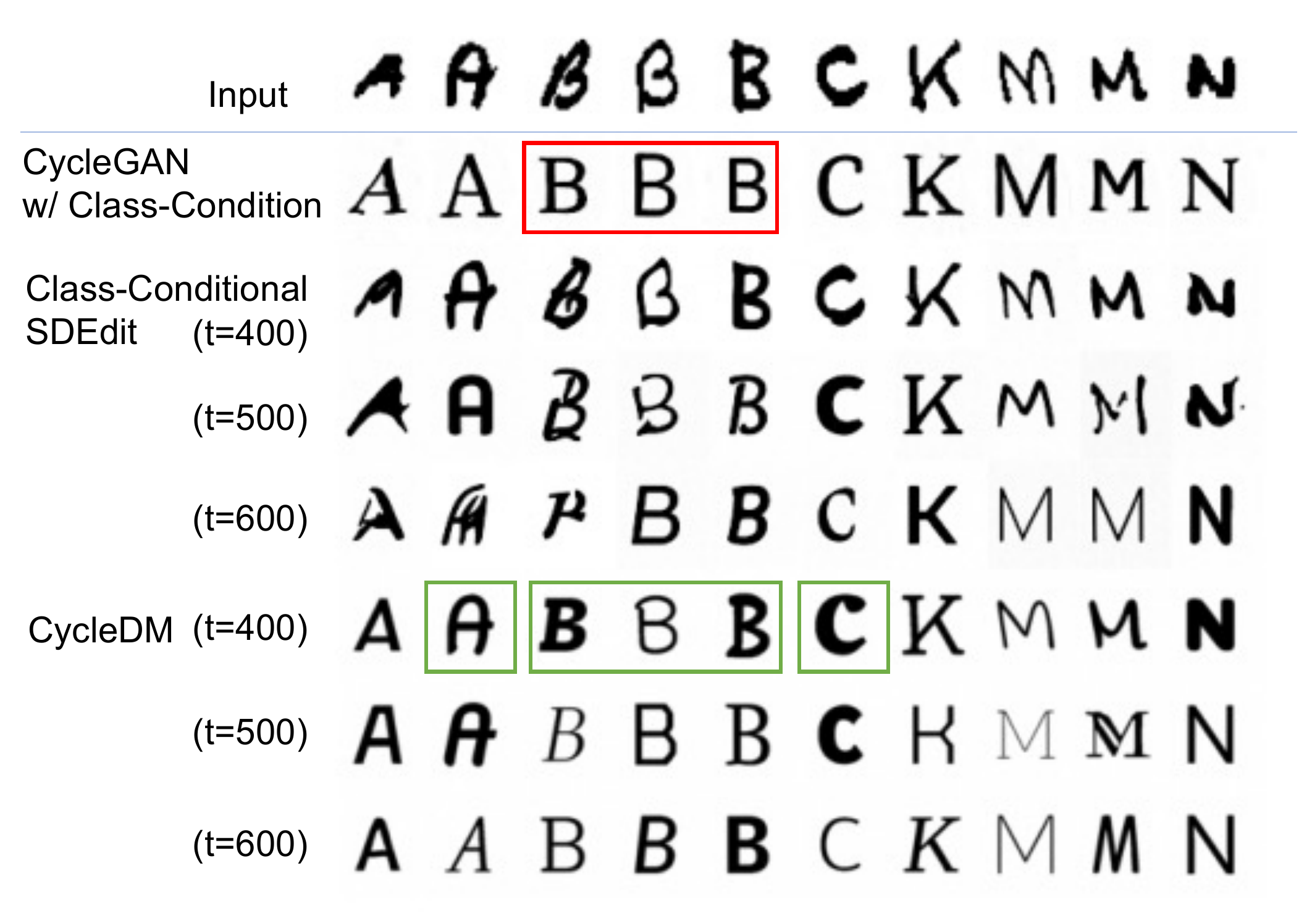}\\[-3mm]
    \caption{Image conversion from the handwritten character domain to the machine-printed character domain (HW$\to$MP). The green boxes are attached to the results subjectively appropriate, whereas the red boxes are inappropriate.}
    \label{fig:hw_to_mp}
\end{figure}%
% 提案法が全体的に活字らしい文字が生成できている（緑枠AとC）
% たとえば，Aでは丸みを帯びた上部の形状を再現できている
% Cはストローク終わりの払いが再現
% 一方で，提案法はさまざまなスタイルで生成できている(緑枠B)
% 提案法でも，tが小さい方がより手書きの特徴を捉えた活字変換ができている
% tが大きいとより活字らしい文字が生成されるが手書きの特徴が落ちる(トレードオフ)
% CycleGANでは似たような文字ばかりを出す傾向（赤枠のB）
% SDEditは手書きの特徴が残りやすい．特にtが小さい時

We conducted qualitative evaluations to observe how the cross-domain conversion was performed in a style-consistent manner. In other words, we expect that the appearance of the original image in the source domain needs to be kept in the converted result in the target domain. By observing the actual conversion pairs, it is possible to confirm whether this expectation is valid or not.
\par
% 12:15 SU

\subsubsection{Conversion from HW to MP}
Fig.~\ref{fig:hw_to_mp} shows the generated MPs from HWs.
CycleDM could convert HWs into MPs while not only keeping the original HWs' characteristics but also removing irregularities in HWs. For example, `A' and `C' highlighted in green boxes have similar shapes to the input HWs. The `A' has an arch shape on the top stroke, and the `C' has a tapered stroke end. At the same time, their curves become smoother, and their stroke widths become more consistent. \par
% 13:15 SU

The observation of the diversity in the generated MPs is important.
As indicated by the variations in `B's shapes, CycleDM could generate MPs with different styles, whereas CycleGAN could not --- it generates similar `B's regardless of the diversity of the HW inputs. Although SDEdit generated MPs in various styles, 
their readability is often not enough; in some examples, the generated MPs are hard to read. These observations coincide with the quantitative evaluation result in Table~\ref{g:mp_result}.
\par
% 13:30 SU
It is also important to observe the effect of $t$ in the results of CycleDM. The smaller $t$ becomes, the more the generated images keep the style of the original HWs; conversely, the larger $t$, the more the generated image looks like MP, and the original HW style is lost.
\par
% 13:34 SU

\begin{figure}[t]
    \centering
    \includegraphics[width=0.8\linewidth]{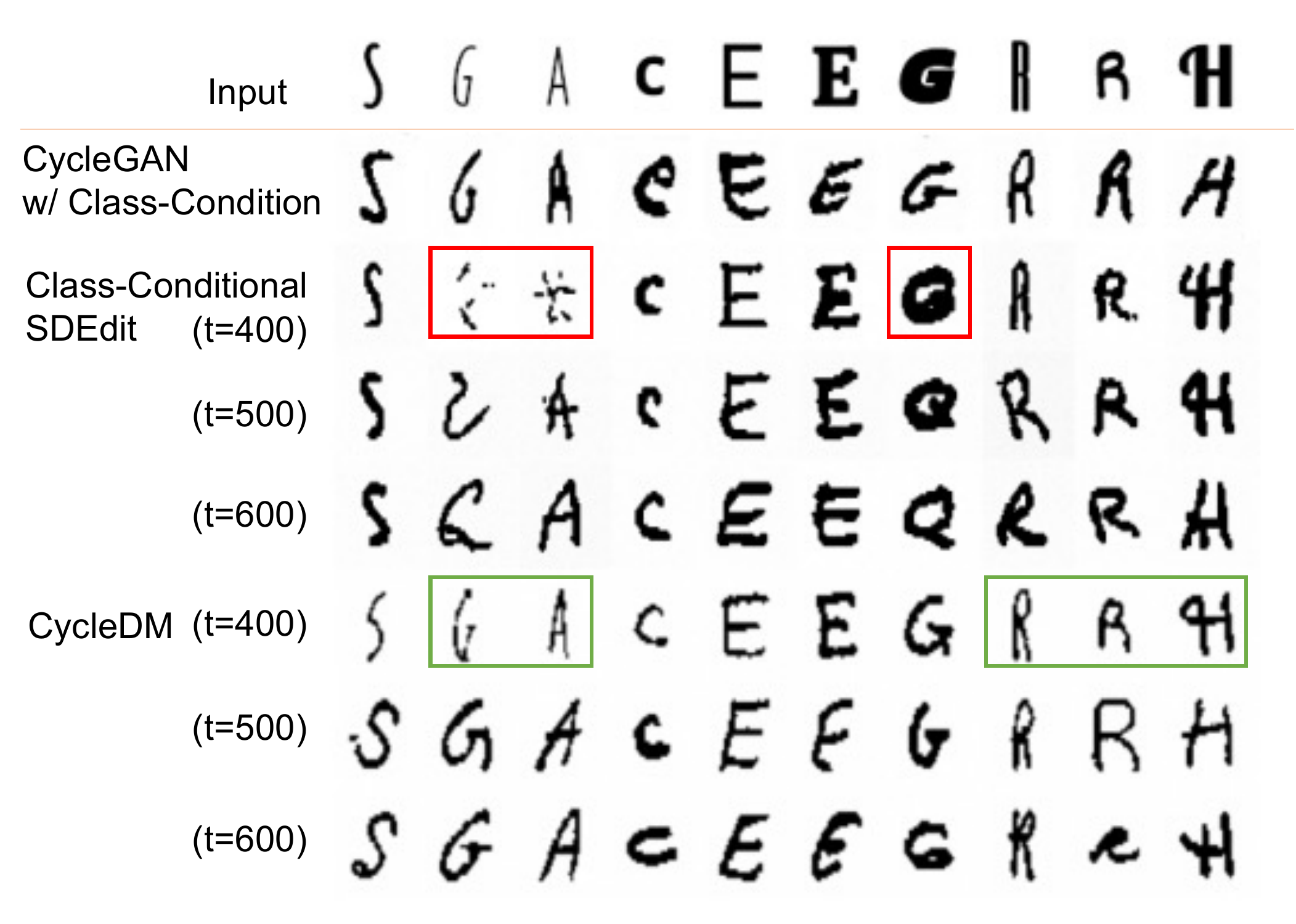}\\[-3mm]
    \caption{Image conversion from the machine-printed character domain to the handwritten character domain (MP$\to$HW).}
    \label{fig:mp_to_hw}
\end{figure}%
% 提案法はいろんな活字(thinやbold,一部では特殊な活字)に対応可能
% 提案法はthinやboldどちらも対応できる． (緑枠CA)
% 特に，提案法で面白い例としてcondencedなRや元々手書き風なフォント？のR
% Hは左上の跳ねているスタイルを表現したまま手書きに変換できている
% ただし，手書き->活字の時同様，tが大きくなるにつれて，手書きの特徴が薄くなっていく
% 例えば，H（緑枠）は最初は活字にも見られる左上のストロークが，tが大きくなるとなくなっている．

% CycleGANは元のフォントの特徴を全然保っていない．
% SDeditは，ストロークの細い文字に弱い．(赤枠G,A)
% また，boldはストロークが潰れてしまうことがある(赤枠G)
% この欠点はおそらく数ピクセルの増減により，thinなフォントはストロークがなくなり
% boldなフォントはストロークの間が埋め尽くされたためと考えられる

\subsubsection{Conversion from MP to HW}
Fig.~\ref{fig:mp_to_hw} shows the generated HWs from real MPs. CycleDM could convert various MPs into HWs, while showing the original style of MPs.
For example, as shown in `G' and `A' in a green box, CycleDM could convert the MPs to HW-like versions with thin strokes. \par
More interestingly, CycleDM could convert decorative MPs to HWs.
The MPs of `R' and `H' have a condensed or fancy style and are converted to HWs showing the same styles (especially when $t=400$). In contrast, the second `E' has a serif at the end of each stroke. However, almost all generated HW does not have serifs. Since serifs are specific to MPs and we do not write them in our HWs, the generated HWs also do not have them.\par
% SU 13:48
SDEdit seems to have a hard problem with MPs with thin strokes, as shown in `G' and `A' in a red box. As the result of adding a large amount of noise at $t=400$, the structure by thin strokes
is destroyed, and it is difficult to generate HW-like images while keeping the original MP styles.
On the other hand, the MP `G' with a heavy stroke is converted into its HW version while losing the details as `G.' This is also because of the large noise that moves the details (of the narrow background).\par
% SU 14:01

\begin{figure}[t]
    \centering
    \includegraphics[width=0.8\linewidth]{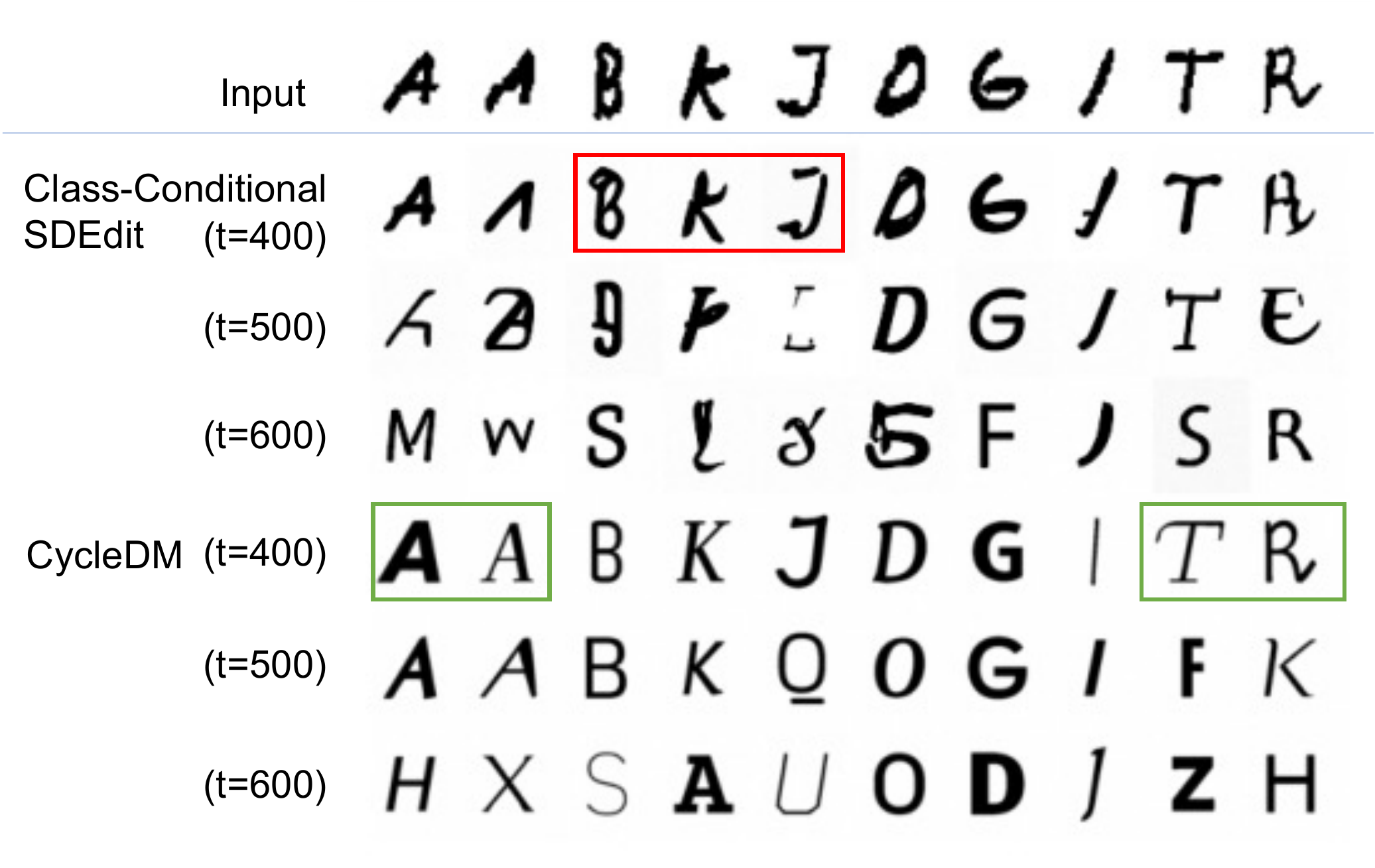}\\[-3mm]
    \caption{Image conversion from the handwritten character domain to the machine-printed character domain (HW$\to$MP) without class condition.}  
    \label{fig:hw_to_mp_uncond}
\end{figure}
% 提案法はt=400はある程度生成可能．tが大きくなると性能下がる．特に，t=600は全然ダメ
% 提案法はinputが違うと生成も結構違う(variationがあってhappy?) (緑枠A)
% 提案法では手書きの特徴を残した面白い生成結果を示している（緑枠T,R）
% Baselineはできている文字もあるが，inputが似ていると同じものだしがち（赤枠A）
% また，baselineは最近傍なので，既知のフォントのみしか出せない (フォント生成の目的から考えると不十分)
% SDEditはかなり厳しい結果．t=400は文字種は結構近いものの，手書きの特徴残りすぎ
% tが大きくなると提案手法同様文字種が全然当たらない
\subsubsection{Is the class condition $c$ important for conversion?}
Finally, Fig.~\ref{fig:hw_to_mp_uncond} shows the converted MPs from HWs without class condition $c$.
When $t=400$, CycleDM could generate MP-like character images that reflect the original HWs. Moreover, we also can observe the variations in `A.' However, when $t=600$, the class information in the original HWs is often lost, and the resulting MPs become characters in a different class. Consequently, we need to be more careful of $t$ when we do not specify the class $c$. Severer results are found with SDEdit.
% Su 14:11
%===================================================================
\section{Conclusion, Limitation, and Future Work}
% ===================================================================．
%本論文では，CycleGANを拡散モデルに導入した新たな画像変換手法を提案し，文字画像の相互変換を行った．具体的には，ノイズ画像同士をCycleGANと同様のアプローチによって変換モデルを学習する．
%そして，学習されたモデルによって変換されたノイズ画像を，学習済みの拡散モデルでデノイズすることにより，高品質な画像変換が可能となった．
%実験結果から，この手法を用いることで，その文字固有の特徴を保持しながら文字画像の変換が可能であることが分かった．

We proposed a novel image conversion model called CycleDM and applied it to cross-modal conversion between handwritten and machine-printed character images. We experimentally proved that CycleDM shows better performance quantitatively and qualitative than SDEdit and CycleGAN, both of which are state-of-the-art image conversion models.
Especially we showed that CycleDM can keep the original style in the conversion results; moreover, CycleDM is useful for converting handwritten character images into machine-printed styles as a preprocessing for OCR.\par

% ---limitation---
%現在の変換モデルは，時刻$t$を固定にしたモデルであり，別な時刻$t$で画像生成を試みる場合，新たな変換モデルの学習が必要になる．拡散モデルの全時刻$t$のノイズ画像に対して学習を行うことでより効率的なモデルが実現される．
%また，今回の変換モデルは，文字クラスをClass-Conditionとして与えて学習を行っており，完全な条件なしの変換モデルを用いた画像生成ではない．そのため，今後の研究として文字クラスなどのConditionを与えず，画像変換を行うことが挙げられる．
One limitation is that CycleDM performs its conversion at a prefixed time $t$. Although the experimental results show the robustness of CycleDM to $t$ as long as we give the class condition, we can treat $t$ in a more flexible way. Application to non-character images is another possible future work.

% =========================================================================
\par
\noindent{\bf Acknowledgment}:\ This work was supported by JSPS KAKENHI Grant Number JP22H00540.
% =========================================================================
%
\bibliographystyle{splncs04}
\bibliography{sample-base}

\begin{thebibliography}{10}
\providecommand{\url}[1]{\texttt{#1}}
\providecommand{\urlprefix}{URL }
\providecommand{\doi}[1]{https://doi.org/#1}

\bibitem{ahn2023dreamstyler}
Ahn, N., Lee, J., Lee, C., Kim, K., Kim, D., Nam, S.H., Hong, K.: {DreamStyler: Paint by Style Inversion with Text-to-Image Diffusion Models}. arXiv preprint arXiv:2309.06933  (2023)

\bibitem{controlstyle}
Chen, J., Pan, Y., Yao, T., Mei, T.: {ControlStyle: Text-Driven Stylized Image Generation Using Diffusion Priors}. In: ACM International Conference on Multimedia. pp. 7540--–7548 (2023)

\bibitem{cohen2017emnist}
Cohen, G., Afshar, S., Tapson, J., Van~Schaik, A.: Emnist: Extending mnist to handwritten letters. In: 2017 international joint conference on neural networks (IJCNN). pp. 2921--2926. IEEE (2017)

\bibitem{dhariwal2021diffusion}
Dhariwal, P., Nichol, A.: {Diffusion Models Beat GANs on Image Synthesis}. In: Advances in Neural Information Processing Systems (NeurIPS). pp. 8780--8794 (2021)

\bibitem{goodfellow2014gan}
Goodfellow, I.J., Pouget-Abadie, J., Mirza, M., Xu, B., Warde-Farley, D., Ozair, S., Courville, A., Bengio, Y.: {Generative Adversarial Nets}. In: Advances in Neural Information Processing Systems (NeurIPS) (2014)

\bibitem{gui2023zero}
Gui, D., Chen, K., Ding, H., Huo, Q.: {Zero-shot Generation of Training Data with Denoising Diffusion Probabilistic Model for Handwritten Chinese Character Recognition}. In: International Conference on Document Analysis and Recognition. p. 348–365 (2023)

\bibitem{ishaan2017wgan}
Gulrajani, I., Ahmed, F., Arjovsky, M., Dumoulin, V., Courville, A.: {Improved Training of Wasserstein GANs}. In: Advances in Neural Information Processing Systems (NeurIPS) (2017)

\bibitem{Hamazaspyan_2023_CVPR}
Hamazaspyan, M., Navasardyan, S.: {Diffusion-Enhanced PatchMatch: A Framework for Arbitrary Style Transfer With Diffusion Models}. In: The IEEE/CVF Conference on Computer Vision and Pattern Recognition (CVPR) Workshops. pp. 797--805 (2023)

\bibitem{he2022diff}
He, H., Chen, X., Wang, C., Liu, J., Du, B., Tao, D., Qiao, Y.: {Diff-Font: Diffusion Model for Robust One-Shot Font Generation}. arXiv preprint arXiv:2212.05895  (2022)

\bibitem{he2023wordart}
He, J.Y., Cheng, Z.Q., Li, C., Sun, J., Xiang, W., Lin, X., Kang, X., Jin, Z., Hu, Y., Luo, B., et~al.: {WordArt Designer: User-Driven Artistic Typography Synthesis using Large Language Models}. In: Proceedings of the 2023 Conference on Empirical Methods in Natural Language Processing: Industry Track (EMNLP). pp. 223--232 (2023)

\bibitem{FID}
Heusel, M., Ramsauer, H., Unterthiner, T., Nessler, B., Hochreiter, S.: {GANs Trained by a Two Time-Scale Update Rule Converge to a Local Nash Equilibrium}. In: Advances in Neural Information Processing Systems(NeurIPS) (2017)

\bibitem{ho2020denoising}
Ho, J., Jain, A., Abbeel, P.: {Denoising Diffusion Probabilistic Models}. In: Advances in Neural Information Processing Systems (NeurIPS) (2020)

\bibitem{DiffStyler}
Huang, N., Zhang, Y., Tang, F., Ma, C., Huang, H., Dong, W., Xu, C.: {DiffStyler: Controllable Dual Diffusion for Text-Driven Image Stylization}. IEEE Transactions on Neural Networks and Learning Systems pp. 1--14 (2024)

\bibitem{ide2017does}
Ide, S., Uchida, S.: {How Does a CNN Manage Different Printing Types?} In: Proceedings of the 14th IAPR International Conference on Document Analysis and Recognition (ICDAR). pp. 1004--1009 (2017)

\bibitem{IluzVinker2023}
Iluz, S., Vinker, Y., Hertz, A., Berio, D., Cohen-Or, D., Shamir, A.: {Word-As-Image for Semantic Typography}. ACM Transactions on Graphics  \textbf{42}(4) (2023)

\bibitem{isola2017pix2pix}
Isola, P., Zhu, J.Y., Zhou, T., Efros, A.A.: {Image-to-Image Translation with Conditional Adversarial Networks}. In: Proceedings of the IEEE Conference on Computer Vision and Pattern Recognition (CVPR). pp. 1125--1134 (2017)

\bibitem{kynkaanniemi2019improved}
Kynk{\"a}{\"a}nniemi, T., Karras, T., Laine, S., Lehtinen, J., Aila, T.: Improved precision and recall metric for assessing generative models. In: Advances in Neural Information Processing Systems(NeurIPS). vol.~32 (2019)

\bibitem{meng2022sdedit}
Meng, C., He, Y., Song, Y., Song, J., Wu, J., Zhu, J.Y., Ermon, S.: {SDEdit: Guided Image Synthesis}. In: Proceedings of The International Conference on Learning Representations (ICLR) (2022)

\bibitem{murez2018image}
Murez, Z., Kolouri, S., Kriegman, D., Ramamoorthi, R., Kim, K.: {Image to Image Translation for Domain Adaptation}. In: Proceedings of the IEEE Conference on Computer Vision and Pattern Recognition (CVPR). pp. 4500--4509 (2018)

\bibitem{Pan_2023_WACV}
Pan, Z., Zhou, X., Tian, H.: {Arbitrary Style Guidance for Enhanced Diffusion-Based Text-to-Image Generation}. In: The IEEE/CVF Winter Conference on Applications of Computer Vision (WACV). pp. 4461--4471 (2023)

\bibitem{Rombach_2022_CVPR}
Rombach, R., Blattmann, A., Lorenz, D., Esser, P., Ommer, B.: {High-Resolution Image Synthesis With Latent Diffusion Models}. In: Proceedings of the IEEE/CVF Conference on Computer Vision and Pattern Recognition (CVPR). pp. 10684--10695 (June 2022)

\bibitem{unet}
Ronneberger, O., Fischer, P., Brox, T.: {U-Net: Convolutional Networks for Biomedical Image Segmentation}. In: Proceedings of the International Conference on Medical Image Computing and Computer Assisted Intervention (MICCAI). pp. 234--241 (2015)

\bibitem{palette}
Saharia, C., Chan, W., Chang, H., Lee, C., Ho, J., Salimans, T., Fleet, D., Norouzi, M.: {Palette: Image-to-Image Diffusion Models}. In: Special Interest Group on Computer Graphics and Interactive Techniques Conference (ACM SIGGRAPH). pp. 1--10 (2022)

\bibitem{sasaki2021unit}
Sasaki, H., Willcocks, C.G., Breckon, T.P.: {UNIT-DDPM: UNpaired Image Translation with Denoising Diffusion Probabilistic Models}. arXiv preprint arXiv:2104.05358  (2021)

\bibitem{staindiff}
Shen, Yiqingand~Ke, J.: {StainDiff: Transfer Stain Styles of Histology Images with Denoising Diffusion Probabilistic Models and Self-ensemble}. In: The Medical Image Computing and Computer Assisted Intervention (MICCAI). pp. 549--559 (2023)

\bibitem{shirakawa2023}
Shirakawa, T., Uchida, S.: {Ambigram Generation by a Diffusion Model}. In: Proceedings of 17th International Conference on Document Analysis and Recognition (ICDAR). pp. 314--330 (2023)

\bibitem{SGDiff}
Sun, Z., Zhou, Y., He, H., Mok, P.: {SGDiff: A Style Guided Diffusion Model for Fashion Synthesis}. In: ACM International Conference on Multimedia. pp. 8433--8442 (2023)

\bibitem{Tanveer_2023_ICCV}
Tanveer, M., Wang, Y., Mahdavi-Amiri, A., Zhang, H.: {DS-Fusion: Artistic Typography via Discriminated and Stylized Diffusion}. In: Proceedings of the IEEE/CVF International Conference on Computer Vision (ICCV). pp. 374--384 (2023)

\bibitem{Tzeng_2017_CVPR}
Tzeng, E., Hoffman, J., Saenko, K., Darrell, T.: {Adversarial Discriminative Domain Adaptation}. In: Proceedings of the IEEE Conference on Computer Vision and Pattern Recognition (CVPR). pp. 7167--7176 (2017)

\bibitem{uchida2016further}
Uchida, S., Ide, S., Iwana, B.K., Zhu, A.: {A Further Step to Perfect Accuracy by Training CNN with Larger Data}. In: 2016 15th International Conference on Frontiers in Handwriting Recognition (ICFHR). pp. 405--410 (2016)

\bibitem{wang2023anything}
Wang, C., Wu, L., Liu, X., Li, X., Meng, L., Meng, X.: {Anything to Glyph: Artistic Font Synthesis via Text-to-Image Diffusion Model}. In: SIGGRAPH Asia 2023 Conference Papers. pp. 1--11 (2023)

\bibitem{wang2022semantic}
Wang, W., Bao, J., Zhou, W., Chen, D., Chen, D., Yuan, L., Li, H.: Semantic image synthesis via diffusion models. In: arXiv preprint arXiv:2207.00050 (2022)

\bibitem{Wu_2023_ICCV}
Wu, C.H., De~la Torre, F.: {A Latent Space of Stochastic Diffusion Models for Zero-Shot Image Editing and Guidance}. In: Proceedings of the IEEE/CVF International Conference on Computer Vision (ICCV). pp. 7378--7387 (2023)

\bibitem{yang2023zero}
Yang, S., Hwang, H., Ye, J.C.: {Zero-Shot Contrastive Loss for Text-Guided Diffusion Image Style Transfer}. arXiv preprint arXiv:2303.08622  (2023)

\bibitem{yang2024fontdiffuser}
Yang, Z., Peng, D., Kong, Y., Zhang, Y., Yao, C., Jin, L.: {FontDiffuser: One-Shot Font Generation via Denoising Diffusion with Multi-Scale Content Aggregation and Style Contrastive Learning}. In: Proceedings of the AAAI conference on artificial intelligence (2024)

\bibitem{Zhang_2023_ICCV}
Zhang, L., Rao, A., Agrawala, M.: {Adding Conditional Control to Text-to-Image Diffusion Models}. In: Proceedings of the IEEE/CVF International Conference on Computer Vision (ICCV). pp. 3836--3847 (2023)

\bibitem{zhu2020cyclegan}
Zhu, J.Y., Park, T., Isola, P., Efros, A.A.: {Unpaired Image-to-Image Translation Using Cycle-Consistent Adversarial Networks}. In: Proceedings of the Conference on Computer Vision and Pattern Recognition (CVPR). pp. 2223--2232 (2017)

\end{thebibliography}

\end{document}